\documentclass[a4paper,twoside]{article}
\pdfoutput=1
\usepackage{epsfig}
\usepackage{subcaption}
\usepackage{calc}
\usepackage{amssymb}
\usepackage{amstext}
\usepackage{amsmath}
\usepackage{amsthm}
\usepackage{multicol}
\usepackage{pslatex}
\usepackage{apalike}
\usepackage{algorithm2e}
\usepackage[bottom]{footmisc}
\usepackage{comment}
\usepackage{SCITEPRESS}     

\begin{document}

\title{M\&M: Multimodal-Multitask Model Integrating Audiovisual Cues in Cognitive Load Assessment}

\author{\authorname{Long Nguyen-Phuoc\sup{1,2}\orcidAuthor{0000-0003-1294-8360}, Renald~Gaboriau\sup{2}\orcidAuthor{0000-0001-5565-5088}, Dimitri~Delacroix\sup{2}\orcidAuthor{0009-0002-2143-7343} and Laurent~Navarro\sup{1}\orcidAuthor{0000-0002-8788-8027}}
\affiliation{\sup{1}Mines Saint-Étienne, University of Lyon, University Jean Monnet, Inserm, U 1059 Sainbiose, Centre CIS, 42023 Saint-Étienne, France}
\affiliation{\sup{2}MJ Lab, MJ INNOV, 42000 Saint-Etienne, France}
\email{\{long.nguyen-phuoc, navarro\}@emse.fr, \{renald.gaboriau, dimitri.delacroix\}@mjinnov.com}
}


\abstract{This paper introduces the M\&M model, a novel multimodal-multitask learning framework, applied to the AVCAffe dataset for cognitive load assessment (CLA). M\&M uniquely integrates audiovisual cues through a dual-pathway architecture, featuring specialized streams for audio and video inputs. A key innovation lies in its cross-modality multihead attention mechanism, fusing the different modalities for synchronized multitasking. Another notable feature is the model's three specialized branches, each tailored to a specific cognitive load label, enabling nuanced, task-specific analysis. While it shows modest performance compared to the AVCAffe's single-task baseline, M\&M demonstrates a promising framework for integrated multimodal processing. This work paves the way for future enhancements in multimodal-multitask learning systems, emphasizing the fusion of diverse data types for complex task handling.}

\onecolumn \maketitle \normalsize \setcounter{footnote}{0} \vfill

\section{\uppercase{Introduction}}
\label{sec:introduction}

In the dynamic field of cognitive load assessment (CLA), understanding complex mental states is crucial in diverse areas like education, user interface design, and mental health. Cognitive load refers to the effort used to process information or to perform a task and can vary depending on the complexity of the task and the individual's ability to handle information \cite{Block2010}. In educational psychology, cognitive load, essential for instructional design, reflects the mental demands of tasks on learners \cite{Paas2020}. It suggests avoiding working memory overload for effective learning \cite{Young2014}. In user interface design, it involves minimizing user's mental effort for efficient interaction \cite{NielsenNorman2013}. In mental health, it relates to cognitive tasks' mental workload, significantly impacting those with mental health disorders or dementia \cite{Beecham2017}.

CLA encompasses diverse traditional methods. Dual-Task Methodology in multimedia learning and Subjective Rating Scales in education evaluate cognitive load, requiring validation in complex contexts \cite{Thees2021}. Wearable devices and Passive Brain-Computer Interfaces offer real-time and continuous monitoring \cite{Jaiswal2021,Gerjets2014}. Hemodynamic Response Analysis and linguistic behavioral analysis provide accurate assessment in specific tasks \cite{Ghosh2019,Khawaja2014}. Mobile EEG and physiological data analysis further contribute to multimodal measurement strategies \cite{Kutafina2021,Vanneste2020}.

Recent progress in assessing cognitive load through machine learning and deep learning has been notable. Neural networks, particularly under deep learning frameworks, have achieved accuracies up to 99\%, with artificial neural networks and support vector machines being key techniques \cite{Elkin2017}. Deep learning, especially models like stacked denoising autoencoders and multilayer perceptrons, have outperformed traditional classifiers in estimating cognitive load \cite{Saha2018}. Enhanced methods have shown impressive results in classifying mental load using EEG data, comparing favorably with CNNs \cite{Jiao2018,Kuanar2018Cognitive} and RNN \cite{Kuanar2018Cognitive}. Additionally, machine learning has been effective in detecting cognitive load states using signals like ECG and EMG \cite{OschliesStrobel2017} or PPG \cite{Zhang2019Photoplethysmogram-based}. Although these study highlight the growing role of machine learning in accurately assessing cognitive load, they often fall short in capturing the multifaceted nature of cognitive load, necessitating multimodal learning. 

Our M\&M model is designed to accurately assess cognitive load across multiple contexts, effectively leveraging multimodal inputs and multitask learning. Recognizing the intricate nature of cognitive load, we integrate audiovisual data to capture a comprehensive picture of cognitive state. The primary contributions of our work are:

\begin{itemize}
  \item Incorporation of multimodal data fusion: By leveraging the complementary of audio and video data that mirrors human sensory observation, M\&M ensures a comprehensive and non-intrusive capture of cognitive load.
  \item Implementation of multitask learning: This approach not only simplifies the training process by jointly learning various aspects of cognitive load but also improves the overall accuracy and robustness of the model. 
\end{itemize}

\begin{figure*}[!h]
  \centering
   \includegraphics[width=300pt]{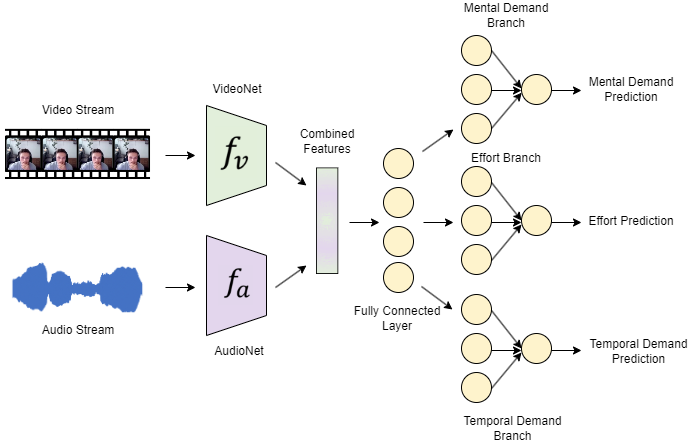}
  \caption{The M\&M model's architecture}
  \label{fig:archi}
 \end{figure*}

\section{\uppercase{Related Works}}

\subsection{Multimodal Learning for CLA}
The field of CLA has witnessed significant advancements through the creation of multimodal datasets and dedicated research in multimodal learning.

Dataset creation initiatives \cite{Mijic2019,Gjoreski2020,Oppelt2022,sarkar2022avcaffe} prioritize the development of robust, annotated datasets that capture a range of modalities, including physiological signals, facial expressions, and environmental context. These datasets serve as critical benchmarks for evaluating and training CLA models, ensuring that researchers have access to high-quality, diverse data sources for algorithm development.

In contrast, studies focused on multimodal learning in CLA leverage existing datasets to investigate and refine techniques for integrating and interpreting data from multiple sources. \cite{Chen2020} estimates task load levels and types from eye activity, speech, and head movement data in various tasks using event intensity, duration-based features, and coordination-based event features. \cite{Cardone2022} evaluates drivers' mental workload levels using machine learning methods based on ECG and infrared thermal signals, advancing traffic accident prevention. \cite{Daza2023MATT:} introduces a multimodal system using Convolutional Neural Networks to estimate attention levels in e-learning sessions by analyzing facial gestures and user actions.

\subsection{Multitask Learning for CLA}
Recent advancements in multitask learning (MTL) neural networks could have significantly impacted CLA. These advancements include: improved methods \cite{Ruder2017}, personalized techniques \cite{Taylor2020}, data efficiency and regularization \cite{Sogaard2017}.

Moreover, cutting-edge developments, such as the introduction of adversarial MTL neural networks, have further propelled this field. These networks are designed to autonomously learn task relation coefficients along with neural network parameters, a feature highlighted in \cite{Zhou2020}. However, despite these advancements, \cite{Gjoreski2020} remains one of the few studies demonstrating the superiority of MTL networks over single-task networks in the realm of CLA. This indicates the potential yet unexplored in fully harnessing the capabilities of MTL in complex and nuanced areas like cognitive load assessment.

\subsection{Multimodal-Multitask Learning for CLA}
In our exploration of the literature, we found no specific studies on Multimodal-Multitask Neural Networks tailored for cognitive load assessment. This gap presents an opportunity for pioneering research. Therefore, in this section, we broaden our focus to  the wider domain of cognitive assessment.

\cite{Tan2021} developed a bioinspired multisensory neural network capable of sensing, processing, and memorizing multimodal information. This network facilitates crossmodal integration, recognition, and imagination, offering a novel approach to cognitive assessment by leveraging multiple sensory inputs and outputs. \cite{ElSappagh2020} presented a multimodal multitask deep learning model specifically designed for detecting the progression of Alzheimer's disease using time series data. This model combines stacked convolutional neural networks and BiLSTM networks, showcasing the effectiveness of multimodal multitask approaches in tracking complex mental health conditions. \cite{Qureshi2019} demonstrated a network that improves performance in estimating depression levels through multitask representation learning. By fusing all modalities, this network outperforms single-task networks, highlighting the benefits of integrating multiple data sources for accurate cognitive assessment.

\section{\uppercase{Methods}}
Our motivation for developing the M\&M model stems from several key objectives. Firstly, it aims to offer a compact and efficient AI solution, easing training demands in settings with limited computational power. In the context of human-robot interaction, M\&M simplifies deployment by unifying multiple labels and inputs into one model, reducing the infrastructure needed for effective operation. Lastly, it confirms the interest of multimodality for all studies related to the analysis of information provided by human beings \cite{kress_multimodality_2009}.

\subsection{Signal Processing}
\subsubsection{Audio Processing} \label{audioproc}
The preprocessing stage consist first in downsampling the audio stream at 16 kHz. The waveform is then zero-padded or truncated to match the same length. Afterward, the mel-spectrogram \cite{AriasVergara2021} of the segments are computed using \( n_{\text{mels}} \) = 80 mel filters with a Fourier transform window size of 1024 points. The hop length, which determines the stride for the sliding window, is set to 1\% of the target sample rate, effectively creating a 10 ms hop size between windows. The MelSpectrogram transformation is particularly suitable for audio processing because it mirrors human auditory perception more closely than a standard spectrogram. 

\subsubsection{Video Processing} \label{videoproc}
The visual stream is downsampled at 5 frames per second and resized to a spatial resolution of of \( 168 \times 224 \) pixels, reducing computational load for faster training. Zero padding and truncating are also needed to obtain same length. After this step, all frames are transformed in the following pipeline. Initially, video frames are randomly flipped horizontally half the time, imitating real-world variations. A central portion of each frame is then cropped, sharpening the focus on key visual elements. Converting frames to grayscale emphasizes structural details over color, reducing the model's processing load and bias due to color temperatures of different videos. Finally, normalization, with means and standard deviations both set to 0.5, ensures the pixel values to be in the comparable range of [0, 1] which stabilizes training and reduces skewness.

\subsection{Model Architecture}

 \begin{figure*}[h]
\centering
\footnotesize
\begin{tabular}{l c c c c c c}
&  & \textbf{Participant A} &  &  & \textbf{Participant B} &  \\
 & Open discussion & Montclair map & Multi-task & Open discussion & Montclair map & Multi-task \\
\multicolumn{1}{c}{} & \multicolumn{3}{c}{\includegraphics[width=170pt]{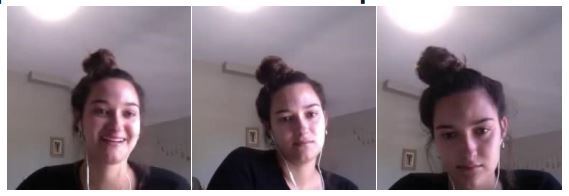}} & \multicolumn{3}{c}{\includegraphics[width=170pt]{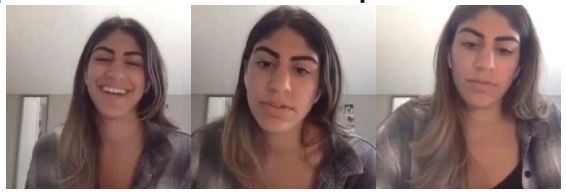}} \\
\textbf{Effort} & 8 & 17 & 16 & 2 & 16 & 12 \\
\textbf{Mental demand} & 3 & 12 & 13 & 4 & 12 & 15 \\
\textbf{Temporal demand} & 0 & 20 & 11 & 3 & 16 & 6 \\
\end{tabular}
\caption{Some examples of the AVCAffe dataset. The self-reported cognitive score shown here at the NASA-TLX scale.}
\end{figure*}

The M\&M's architecture, illustrated in Figure \ref{fig:archi}, can be segmented into four principal components, each contributing uniquely to the overall functionality of the model.

\subsubsection{AudioNet}
AudioNet, a convolutional neural network architecture \cite{wang2023cnns} designed for processing audio data, accepts as input a Mel spectrogram, represented by the tuple \textit{\( (n_{\text{mels}}, seq_a) \)}. Here, \( n_{\text{mels}} \) denotes the number of mel filters, and \( seq_a \)
is the length of the audio tensor obtained after the transformation of raw audio input as described in Section \ref{audioproc}. Precisely, AudioNet consists of four convolutional layers, each followed by batch normalization. The convolutional layers progressively increase the number of filters from 16 to 128, extracting hierarchical features from the audio input. A max pooling layer follows the convolutional blocks to reduce the spatial dimensions of the feature maps. The network includes a dropout layer set at 0.5 to prevent overfitting. The flattened output from the convolutional layers is then fed into a fully connected layer with 128 units. The AudioNet uses ReLU activation functions and is capable of end-to-end training, transforming raw audio input into a compact representation.

\subsubsection{VideoNet}
VideoNet in the M\&M model, based on the I3D backbone \cite{carreira2018quo}, is a sophisticated neural network tailored for visual data processing. It commences with sequential 3D convolutional layers, succeeded by max-pooling layers, instrumental in feature extraction and dimensionality reduction. The network's essence lies in its Inception modules, adept at capturing multi-scale features. These modules amalgamate distinct convolutional and pooling branches, facilitating a comprehensive feature analysis. Post-feature extraction, an adaptive average pooling layer condenses the data, culminating in a fully connected layer with 128 units after dropout and linear transformation. The network takes as input video data represented by the tuple \textit{\( (\text{d}, \text{h}, \text{w}) \)}, where \textit{\( \text{h}\)} and \textit{\( \text{w} \)} reflecting the post-processed video spatial dimensions and \textit{\( \text{d} \)} the depth which is the product of clip length and target frame rate as explained in Section \ref{videoproc}. 

\subsubsection{Crossmodal Attention}
Cross-modality attention is a concept derived from the attention mechanism \cite{phuong2022formal}. It has been adapted for scenarios where the model must attend to and integrate information from multiple different modalities, such as audio, text, and images \cite{tsai_multimodal_2019}. Consider two modalities, Audio and Video, the cross-modality attention can be represented as:

\begin{equation}\label{eq:cross_modality_attention}
\text{Attention}(Q_{\text{v}}, K_{\text{a}}, V_{\text{a}}) = \text{softmax}\left(\frac{Q_{\text{v}}K_{\text{a}}^T}{\sqrt{d_k}}\right)V_{\text{a}}
\end{equation}

In this structure, the attention mechanism takes video features as the query \( Q_v\) and audio features as both key \( K_a\) and value \( V_a\). The attention module inputs features from both audio and video modalities, each with a dimension of 128. It outputs a combined feature set, effectively integrating relevant features from both sources for enhanced processing.

\subsubsection{Multitask Separated Branches}
After processed by a common fully connected layer, the combined features \textit{\( \hat{AV} \)} is distributed to three distinct branches, each tailored to a specific cognitive load task. These branches employ individual sigmoid activation functions in their output layers to predict binary classifications for each task, facilitating a probabilistic interpretation of the model's predictions.

The flexibility to adjust weights or modify the architecture asymmetrically for each branch permits the model to be customized based on the complexity or priority of tasks, hence optimizing performance and ensuring tailored learning \cite{Nguyen2020Clinical}.

However, in a multi-branch neural network with shared layers, the learning dynamics in one branch can significantly influence those in others. This interdependence enhances the model's overall learning capability, as advancements in one branch may depend on simultaneous updates in others. Such interconnection not only highlights the collaborative nature of the learning process across different branches but also fosters a richer information exchange \cite{Fukuda2018Cross-Connected}. This synergy can lead to more robust and comprehensive learning outcomes, leveraging shared insights for improved performance in each task-specific branch.

\subsection{Implementation Details}
The central M\&M architecture leverages gradients from three specialized branches—each corresponding to a unique cognitive load task—to train shared neural network layers. This holistic training approach is facilitated by the Adam optimizer for efficient stochastic gradient descent. The model built with PyTorch's fundamental components, undergoes an end-to-end training process, allowing simultaneous learning across the diverse modalities of audio and visual data, ensuring distinct tasks of CLA.

The Binary Cross-Entropy (BCE) Loss function is utilized individually for each branch to cater to the binary classification nature of our tasks. The BCE loss for a single task is expressed as:

\begin{equation}\label{eq1}
L_{bce} = -\frac{1}{N} \sum_{i=1}^{N} [y_i \cdot \log(p_i) + (1 - y_i) \cdot \log(1 - p_i)]
\end{equation}

The global loss is a weighted sum, allowing us to prioritize tasks asymmetrically based on their complexity or importance:

\begin{equation}\label{eq2}
L_{global} = \sum_{k=1}^{K} w_k \cdot L_{bce}^k
\end{equation}

In these equations, \(N\) is the number of samples, \(y_i\) is the true label, \(p_i\) is the predicted probability for the \(i^{th}\) sample, \(K\) is the total number of tasks, \(w_k\) is the weight for the \(k^{th}\) task, and \(L_{bce}^k\) is the BCE loss for the \(k^{th}\) task. The weights \(w_k\) are adjustable to focus the model's learning on specific tasks based on their difficulty or relevance.

Furthermore, the cross-modal attention mechanism is implemented as explicitly described in Algorithm \ref{algo}, ensuring a cohesive integration of audio and video data streams.

\begin{algorithm}[!h]
 \caption{Cross-Modal Multihead Attention Mechanism.}\label{algo}
 \KwData{Audio features \(A\), Video features \(V\)}
 \KwResult{Combined features set \( \hat{AV} \)}
 \textbf{Hyperparameters:} \(H\), number of attention heads \\
 \textbf{Parameters:} \(W^Q, W^K, W^V, W^O\) weight matrices for query, key, value, and output projections. \\
 \For{\(h \in \{1 \ldots H\}\)}{
  \( Q_h = A W^Q_h \) \;
  \( K_h = V W^K_h \) \;
  \( V_h = V W^V_h \) \;
  \tcp{Compute scaled dot-product attention}
  \( \text{head}_h = \text{Attention}(Q_h, K_h, V_h) \)\; 
  \(\hat{AV} = \text{Concat}(\text{head}_1, \ldots, \text{head}_H) W^O \)
 }
\end{algorithm}

\section{\uppercase{Experiment}}
In this section, extensive experimentation illustrates
the compact and novel character of M\&M. We first introduce the AVCAffe dataset
Section 4.1. Then, the experiment setup, including
the criterion metrics and the hyper-parameters setting
situation, is presented in Section 4.2. In Section
4.3, we compare with the dataset's authors baseline model.

\subsection{AVCAffe}
The AVCAffe dataset \cite{sarkar2022avcaffe} offers rich audio-visual data to study cognitive load in remote work, with 108 hours of footage from a globally diverse participant group. It explores the cognitive impact of remote collaboration through task-oriented video conferencing, employing NASA-TLX for multidimensional cognitive load measurement. Challenges in detailed classification led to the adoption of a binary approach. For model validation and to avoid data leakage, we ensured representative splits and attempted to emulate the authors' data augmentation methods as detailed in Table \ref{dataaug}, despite the lack of public implementations.

\begin{table}[h]
\caption{The parameter details of audio-visual augmentations.}\label{dataaug} \centering
\begin{tabular}{|l|l|}
\hline
\multicolumn{2}{|c|}{\textbf{Audio Augmentation}} \\
\hline
Volume Jitter & range=$\pm$0.2 \\
Time Mask & max size=50, num=2 \\
Frequency Mask & max size=50, num=2 \\
Random Crop & range=[0.6,1.5], \\
 & crop scale=[1.0,1.5] \\
\hline
\multicolumn{2}{|c|}{\textbf{Visual Augmentation}} \\
\hline
Multi-scale Crop & min area=0.2 \\
Horizontal Flip & p=0.5 \\
Color Jitter & b=1.0, c=1.0, s=1.0, h=0.5 \\
Gray Scale & p=0.2 \\
Cutout & max size=50, num=1 \\
\hline
\end{tabular}
\end{table}

\begin{table*}[h]
\centering
\caption{CLA results (F1-Score) on AVCAffe validation set (Binary Classification)}\label{tab:performance}
\fontsize{9pt}{11pt}\selectfont
\begin{tabular}{|c|c|c|c|c|c|}
\hline
\textbf{Modalities} & \textbf{Architecture} & \textbf{Global F1 score} & \textbf{Mental Demand} & \textbf{Effort}  & \textbf{Temporal Demand} \\
\hline
\multicolumn{6}{|c|}{\cite{sarkar2022avcaffe}} \\
\hline
Audio & ResNet+MLP & -- & 0.61 & 0.55 & 0.55 \\
Video & R(2+1)D+MLP & -- & 0.59 & 0.55 & 0.55 \\
\hline
Audio-Video & Multimodal-Single Task & -- & 0.62 & 0.61 & 0.59 \\
\hline
\multicolumn{6}{|c|}{Ours} \\
\hline
Audio & CNN+MLP+Multitask & 0.58 & 0.56 & 0.58 & 0.56 \\
Video & I3D+MLP+Multitask & 0.48 & 0.44 & 0.49 & 0.48 \\
\hline
Audio-Video & Multimodal-Multitask (M\&M) & 0.59 & 0.60 & 0.61 & 0.54 \\
\hline
\end{tabular}
\end{table*}

\subsection{Setup}

\subsubsection{Criterion Metrics}
Following the metric construction in the \cite{sarkar2022avcaffe} that publushes the AVCAffe dataset, we employ weighted F1-measures for evaluation to counterbalance imbalanced class distribution, a common issue in cognitive datasets \cite{kossaifi2019sewadb,busso2008iemocap}. We also calculates a global micro-averaged F1 score, combining all labels \textit{$\langle$Mental demand, Effort, Temporal demand$\rangle$} to provide an overall assessment of the model's performance across all tasks, considering both precision and recall in a unified metric. The formula of the F1-Scores is shown below:

\begin{equation}\label{eq3}
    \text{Weighted F1-score} = \frac{\sum_{i=1}^{N} w_i \times \frac{2 \times \text{TP}_i}{2 \times \text{TP}_i + \text{FP}_i + \text{FN}_i}}{\sum_{i=1}^{N} w_i}
\end{equation}

\begin{equation*}
    \text{Global Micro-Averaged F1} = 
\end{equation*}

\begin{equation}\label{eq4}
2 \times \frac{\sum_{\text{labels}} \text{TP}_{\text{labels}}}
    {\sum_{\text{labels}} \text{TP}_{\text{labels}} + \frac{1}{2}(\sum_{\text{labels}} \text{FP}_{\text{labels}}
    + \sum_{\text{labels}} \text{FN}_{\text{labels}})}
\end{equation}

In the given equations, \( N \) represents the total number of samples, with \( i \) indexing each sample, and \( w_i \) denoting the weight associated with the \( i^{th} \) sample. \( TP_i \), \( FP_i \), and \( FN_i \) correspond to the True Positives, False Positives, and False Negatives for the \( i^{th} \) sample, respectively. The term 'Labels' is defined as the set of all labels under consideration, specifically \textit{$\langle$Mental demand, Effort, Temporal demand$\rangle$}. \( TP_{\text{labels}} \), \( FP_{\text{labels}} \), and \( FN_{\text{labels}} \) represent the sums of True Positives, False Positives, and False Negatives across these labels.

\subsubsection{Hyperparameters}
In this study, several hyperparameters are meticulously chosen to ensure an optimal balance between training efficiency and model accuracy. The learning rate is set at 0.001, utilizing the Adam optimizer for gradual and precise model updates. A step learning rate scheduler adjusts the learning rate every 10 epochs by a factor of 0.1, assisting more effective convergence and overcoming local minima. Training spans 30 epochs on 1 GPU NVIDIA T4 16GB, providing sufficient iterations for multimodal-multitask learning. Additionally, early stopping is implemented with a patience of 10 epochs to halt training when no improvement in validation loss is observed, ensuring training efficiency and preventing overfitting.

For the crossmodal attention using PyTorch multi-head attention component, the number of head is fixed at 4.

Moreover, due to limited computational resources, we utilized randomly only 25 clips of 6 seconds each per participant for each task, aligning with the dataset's average clip length. Thus, the final input dimension becomes \textit{\( (n_{\text{mels}} = 80, seq_a = 601)\)} for AudioNet and \textit{\( (\text{d} = 30, \text{h} = 148, \text{w} = 144) \)} for VideoNet. This resulted in dataset partitions containing 15,213 clips for training, 3,804 for validation, and 4,474 for testing. These clips are subsequently grouped into batches of 256 for model training and evaluation.

\subsection{Experimental Results and Comparison}
We report the experimental results in Table \ref{tab:performance} which outlines a comparative analysis of different architectures applied to the AVCAffe dataset for binary classification of cognitive load, indicated by F1-Score metrics across three task domains: Mental Demand, Effort, and Temporal Demand. It compares results from \cite{sarkar2022avcaffe} with those obtained from our proposed M\&M model.

Overall, the F1 scores obtained range from 0.44 to 0.62 reflecting the CLA complexity. Low scores for temporal demand suggest that the task is more challenging or the data is less qualitative than the others.

\cite{sarkar2022avcaffe} separate models used to process audio, video or both modalities. The highest score achieved is for Mental Demand (0.62) using a audio-visual late fusion approach in a single-task framework. However, no global F1 score was reported, suggesting that the authors chose to focus on task-specific performance rather than an aggregated metric across all tasks.

The M\&M model shows competitive results, particularly in the Mental Demand and Effort categories, where it outperforms the single-modality approaches and is on par with or slightly below the \cite{sarkar2022avcaffe} multimodal single-task model. However, it appears to have a lower score in the Temporal Demand, because even if it achieves improved overall task-average performance, they may still yield degraded performance on Temporal Demand individual. Such behavior conforms the finding of the literature \cite{Nguyen2020Clinical}.

These results could indicate that while our M\&M model has a balanced performance across tasks and benefits from the integration of audio-visual data, there might be room for optimization, especially in an asymmetric  temporal demand branch. This also suggests that M\&M can provide a more compact and computationally efficient solution without significantly sacrificing performance, which can be particularly advantageous in resource-constrained environments or for real-time applications.

\section{\uppercase{Conclusions}}
\label{sec:conclusion}
In summary, the M\&M model is designed to extract and integrate audio and video features using separate streams, fuse these features using a crossmodal attention mechanism, and then apply this integrated representation to multiple tasks through separate branches. This structure enables the handling of complex multimodal data and supports multitasking learning objectives.

The M\&M model bridges a crucial research gap by intertwining multitask learning with the fusion of audio-visual modalities, reflecting the most instinctive human observation methods. This unique approach to AI development sidesteps reliance on more invasive sensors, favoring a naturalistic interaction style.

Moving forward, our research will expand to experimenting with various affective computing datasets using the proposed M\&M model. This will deepen our understanding of how different tasks interact within a multimodal-multitask framework. A key focus will be the development of a tailored weighted loss function, which will be designed based on the correlations observed between these tasks. This innovative approach aims to refine and optimize the model's performance by aligning it more closely with the nuanced relationships inherent in cognitive processing tasks.

\bibliographystyle{apalike}
{\small
\bibliography{PAPER}}

\end{document}